\documentclass[letterpaper]{article} 
\usepackage{aaai2026}  
\usepackage{times}  
\usepackage{helvet}  
\usepackage{courier}  
\usepackage[hyphens]{url}  
\usepackage{graphicx} 
\usepackage{natbib}  
\usepackage{caption} 
\usepackage{algorithm}
\usepackage[noend]{algpseudocode}

\usepackage{newfloat}
\usepackage{listings}
\DeclareCaptionStyle{ruled}{labelfont=normalfont,labelsep=colon,strut=off} 
\floatstyle{ruled}
\newfloat{listing}{tb}{lst}{}
\floatname{listing}{Listing}
\usepackage{amsmath}
\usepackage{stmaryrd}
\usepackage{bm}
\usepackage{amssymb}
\usepackage{multirow}
\usepackage{subcaption}
\usepackage{amsthm}
\usepackage{ifthen}

\newboolean{withappendix}
\setboolean{withappendix}{true}

\newcommand{\potentialremove}[2]{\ifthenelse{\boolean{withappendix}}{#1}{#2}}%

\DeclareMathOperator*{\argmin}{arg\,min}

\newcommand{\citeauthorandyear}[1]{\citeauthor{#1}~(\citeyear{#1})}

\newcommand{\generatesuccessors}{\ensuremath{\mathtt{GenSucc}(\state)}}
\newcommand{\generatesuccessorswithpropagation}{\ensuremath{\mathtt{GenSuccPropagation}(\state, Primal)}}
\newcommand{\dualboundfunc}{\ensuremath{\mathtt{Dual}(\state)}}

\newcommand{\dualboundfuncinp}[1]{\ensuremath{\mathtt{Dual}(#1)}}
\newcommand{\dualboundcpfuncinp}[1]{\ensuremath{\mathtt{DualCP}(#1)}}
\newcommand{\dominates}[2]{\ensuremath{\mathtt{Dominates}(#1, #2)}}
\newcommand{\targetstate}{\ensuremath{\mathtt{TargetState}()}}
\newcommand{\isbasestate}{\ensuremath{\mathtt{IsBaseState}(\state)}}
\newcommand{\basecost}{\ensuremath{\mathtt{BaseCost}(\state)}}
\newcommand{\createcpmodel}{\ensuremath{\mathtt{CPModel}(\state)}}
\newcommand{\propagate}{\ensuremath{\mathtt{Propagate}(\mathcal{X}, \mathcal{C}, \mathcal{D})}}
\newcommand{\issuccessorinfeasible}{\ensuremath{\mathtt{IsSuccInfeasible}(\state_{\predicate{\tau}}, \mathcal{D}')}}
\newcommand{\isstateinfeasible}{\ensuremath{\mathtt{IsInfeasible}(\state, \mathcal{D}')}}

\newcommand{\wttw}{\ensuremath{1|r_i, \delta_i|\sum w_i T_i}}
\newcommand{\jobs}{\ensuremath{\mathcal{J}}}
\newcommand{\unscheduled}{\ensuremath{\mathcal{U}}}
\newcommand{\current}{\ensuremath{t}}
\newcommand{\tardiness}{\ensuremath{T_i}}

\newcommand{\tasks}{\ensuremath{\mathcal{T}}}
\newcommand{\resources}{\ensuremath{\mathcal{R}}}
\newcommand{\capacity}{\ensuremath{C}}
\newcommand{\precedences}{\ensuremath{\mathcal{P}}}
\newcommand{\horizon}{\ensuremath{H}}
\newcommand{\scheduled}{\ensuremath{S}}

\newcommand{\predecessors}{\ensuremath{Pre}}
\newcommand{\partialschedule}{\ensuremath{PS}}
\newcommand{\makespan}[1]{\ensuremath{\mathtt{Makespan}(#1)}}

\newcommand{\weight}{\ensuremath{\mathtt{W}(\state, \tau)}}

\newcommand{\locations}{\ensuremath{\mathcal{C}}}
\newcommand{\distance}[2]{\ensuremath{c}_{#1#2}}
\newcommand{\unvisited}{\unscheduled}
\newcommand{\currentlocation}{\ensuremath{l}}

\newcommand{\reachable}{\ensuremath{R}}
\newcommand{\minto}[1]{\ensuremath{\mathtt{MinTo}(#1)}}
\newcommand{\minfrom}[1]{\ensuremath{\mathtt{MinFrom}(#1)}}
\newcommand{\violated}{\ensuremath{O}}

\newcommand{\starttimes}{\ensuremath{St}}
\newcommand{\processingtimes}{\ensuremath{P}}
\newcommand{\resourceusages}{\ensuremath{U}}

\newcommand{\constraint}[1]{\ensuremath{\mathtt{#1}}}
\newcommand{\disjunctive}{\constraint{Disjunctive}}
\newcommand{\cumulative}{\constraint{Cumulative}}

\newcommand{\predicate}[1]{\ensuremath{\llbracket #1 \rrbracket}}
\newcommand{\state}{\ensuremath{\mathcal{S}}}
\newcommand{\statewttw}{\ensuremath{\state}}
\newcommand{\statercpsp}{\ensuremath{\state}}
\newcommand{\statetsptw}{\ensuremath{\state}}

\newcommand{\lb}[1]{\ensuremath{\mathtt{LB}(#1)}}
\newcommand{\ub}[1]{\ensuremath{\mathtt{UB}(#1)}}

\newcommand{\valuef}[1]{\ensuremath{V}(#1)}

\newcommand{\Pumpkin}{Pumpkin}

\newcommand{\astar}{\ensuremath{\text{A}^*}}

\title{Domain-Independent Dynamic Programming with Constraint Propagation}
\author{
    Imko Marijnissen\textsuperscript{\rm 1},
    J. Christopher Beck\textsuperscript{\rm 2},
    Emir Demirović\textsuperscript{\rm 1},
    Ryo Kuroiwa\textsuperscript{\rm 3, \rm 4}
}
\affiliations{
    \textsuperscript{\rm 1}Delft University of Technology, The Netherlands\\
    \textsuperscript{\rm 2}University of Toronto, Canada\\
    \textsuperscript{\rm 3}National Institute of Informatics, Japan\\
    \textsuperscript{\rm 4}The Graduate University for Advanced Studies, SOKENDAI, Japan\\
    i.c.w.m.marijnissen@tudelft.nl, jcb@mie.utoronto.ca, e.demirovic@tudelft.nl, kuroiwa@nii.ac.jp
}

\begin{document}

\maketitle

\begin{abstract}
    There are two prevalent model-based paradigms for combinatorial problems: 1) state-based representations, such as heuristic search, dynamic programming (DP), and decision diagrams, and 2) constraint and domain-based representations, such as constraint programming (CP), \mbox{(mixed-)integer} programming, and Boolean satisfiability.
    In this paper, we bridge the gap between the DP and CP paradigms by integrating constraint propagation into DP, enabling a DP solver to prune states and transitions using constraint propagation.
    To this end, we implement constraint propagation using a general-purpose CP solver in the Domain-Independent Dynamic Programming framework and evaluate using heuristic search on three combinatorial optimisation problems: Single Machine Scheduling with Time Windows (\wttw), the Resource-Constrained Project Scheduling Problem (RCPSP), and the Travelling Salesperson Problem with Time Windows (TSPTW).
    Our evaluation shows that constraint propagation significantly reduces the number of state expansions, causing our approach to solve more instances than a DP solver for \wttw{} and RCPSP, and showing similar improvements for tightly constrained TSPTW instances.
    The runtime performance indicates that the benefits of propagation outweigh the overhead for constrained instances, but that further work into reducing propagation overhead could improve performance further.
    Our work is a key step in understanding the value of constraint propagation in DP solvers, providing a model-based approach to integrating DP and CP.
\end{abstract}

\begin{links}
    \link{Code}{https://doi.org/10.5281/zenodo.19051392}
    \link{Experiments}{https://doi.org/10.5281/zenodo.19051248}
    \potentialremove{}{\link{Appendices}{https://arxiv.org/abs/2603.16648}}
\end{links}

\section{Introduction}
    Combinatorial optimisation is key in Artificial Intelligence (AI), encompassing problems such as planning, scheduling, and variants of satisfiability.
    Within this field, Dynamic Programming (DP) and Constraint Programming (CP) provide declarative model-based frameworks such as Domain-Independent Dynamic Programming~\cite{KUROIWA2026104506} and MiniZinc~\cite{DBLP:conf/cp/NethercoteSBBDT07, DBLP:journals/aim/StuckeyFSTF14} that enable users to formulate representations of problems that can then be solved by generic solvers.
    
    While solving the same problem, the techniques used by solvers following these paradigms differ.
        DP solvers are based on state-based problem representations, which facilitate duplicate/dominance detection techniques.
        While certain DP solvers, such as ddo~\cite{gillard:20:ddo} and CODD~\cite{DBLP:conf/ecai/MichelH24} use decision diagram branch-and-bound~\cite{DBLP:journals/informs/BergmanCHH16}, in this work, we focus on DP solvers using heuristic search by \citeauthorandyear{KUROIWA2026104506}.
        In contrast, CP solvers focus on using depth-first search and inference techniques to prune the search space by determining which values \textit{cannot} be assigned to variables in a solution.

    Previous works introduce constraint propagation in DP approaches, but they either only used problem-specific propagation~\cite{DBLP:journals/eor/FontaineDS23,brița2025optimal}, or did not use heuristic search~\cite{DBLP:journals/scheduling/TanakaF12}. Other, related but orthogonal, work has focused on using dominance and caching~\cite{DBLP:journals/constraints/ChuS15,DBLP:conf/cpaior/ChuBS10,DBLP:conf/cp/Smith05} in a CP solver.
    To the best of our knowledge, no work has been done on integrating generic constraint propagation in a model-based DP framework.
    
    To address this gap, we create a framework for integrating constraint propagation from CP with DP modelling and implement our approach using general-purpose CP and DP solvers, providing:
    \begin{enumerate}
        \item
            A dual view of the problem, using a state-based DP view and an integer-based CP view, which provides:
            \begin{itemize}
                \item
                    Dominance and duplicate detection, and heuristic search using the DP view of the problem.
                \item
                    Strong inference techniques for constraints using the CP view of the problem, allowing pruning of states and strengthening of the dual bound.
            \end{itemize}
        \item 
            A generic model-based way of integrating heuristic search with constraint propagation, allowing \textit{any} form of constraint propagation rather than \textit{only} relying on domain-specific inference techniques (though these techniques are supported by our framework as well).
    \end{enumerate}

    We evaluate our framework on three combinatorial problems:
        Single Machine Scheduling with Time Windows (\wttw),
        the Resource-Constrained Project Scheduling Problem (RCPSP),
        and the Travelling Salesperson Problem with Time Windows (TSPTW).
    For these problems, we show that constraint propagation significantly reduces the number of state expansions, enabling our approach to solve more instances than a DP solver alone for \wttw{} and RCPSP, and tightly constrained TSPTW instances. 
    The performance in terms of runtime indicates that the benefits of propagation outweigh the overhead for tightly constrained instances, but that further work into reducing the propagation time is needed.
    In our in-depth analysis, we vary \wttw{} and TSPTW instance constrainedness, demonstrating that propagation performs exceptionally well when instances are highly constrained.
    
    This work provides a key step in ascertaining the usefulness of constraint propagation in DP solvers using heuristic search, allowing us to better understand the complementary strengths and weaknesses of the approaches.

\section{Preliminaries}
    We introduce the necessary concepts related to dynamic programming, the Domain-Independent Dynamic Programming (DIDP) framework and its Rust interface, heuristic search in DIDP, and constraint programming.

    \subsection{Domain-Independent Dynamic Programming}
        Dynamic Programming (DP) is a problem-solving methodology based on state-based representations.
        \textit{Domain-Independent Dynamic Programming (DIDP)} is a model-based framework for solving combinatorial optimisation problems formulated as DP models.
        One of its strengths is that DIDP models can be solved using generic solvers, removing the need to implement specialised algorithms for each DP model.
        It has recently seen success on combinatorial problems, such as single machine scheduling and TSPTW~\cite{KUROIWA2026104506}.

        A DIDP model is a state-transition system where a solution is created by starting with a \textit{target state} $\state^0$ and applying a sequence of transitions until a \textit{base} state is reached.
        A \textit{successor} state $\state_{\predicate{\tau}}$ is created from \state{} by applying a possible transition $\tau \in \mathrm{T}(\state)$ with weight $w_\tau(\state)$.
        The cost of a solution is the sum of transition weights plus the cost of the base state, where \valuef{\state} maps \state{} to its optimal solution cost and the \textit{dual bound} is a lower (upper) bound on \valuef{\state} in minimisation (maximisation) problems.
        Thus, the goal of a DIDP solver is to compute $\valuef{\state^0}$.
        
        While the DIDP framework allows the user to solve generic DP models, it is limited in its expressivity due to the restrictions of its declarative modelling language, Dynamic Programming Problem Description Language (DyPDL).
        The \textit{Rust Programmable Interface for DIDP (RPID)} defines a DIDP model in terms of Rust functions, allowing the user to define more complex dual bounds, state constraints and/or state variables, while also providing a significant speedup compared to DIDP~\cite{DBLP:conf/cp/0002B25}.

        A RPID model uses three interfaces (called \emph{traits} in Rust) to describe a DP model:
        1) \textbf{Dp} - \targetstate{} defines the target state, \isbasestate{} determines when \state{} is a base state with cost \basecost, and \generatesuccessors{} generates a set of successors $\state_{\predicate{\tau}}$ with transition weight $w_\tau(\state)$, 2) \textbf{Dominance} - \dominates{\state}{\state'} determines if \state{} dominates $\state'$ to prevent exploring redundant states and 3) \textbf{Bound} - \dualboundfunc{} returns the dual bound.

        Using these functions, Equation~\eqref{eq:bellman} recursively defines the value \valuef{\state} of a state (for a minimisation problem) as the minimum value of the successor states and the transition weight, where Inequality~\eqref{eq:dominance_rpid} specifies the dominance relationship and Inequality~\eqref{eq:dual_bound_rpid} the dual bound.
        Thus, solving Equation~\eqref{eq:bellman} provides the optimal solution.
        \begin{flalign}
            & \valuef{\state} =
            \begin{cases}
                \basecost \hspace{1.65cm} \text{if } \isbasestate\\
                \min\limits_{(w_\tau(\state),\ \state_{\predicate{\tau}}) \in \generatesuccessors{}} w_\tau(\state) + \valuef{\state_{\predicate{\tau}}} \hspace{0.1cm} \text{else}%
            \end{cases}\label{eq:bellman}\\
            & \valuef{\state} \leq \valuef{\state'} \quad\hphantom{\text{------------------}} \text{if }\dominates{\state} {\state'}\label{eq:dominance_rpid}\\
            & \valuef{\state} \geq \dualboundfunc\label{eq:dual_bound_rpid}
        \end{flalign}
        \paragraph{Heuristic Search}
            Heuristic search is a strategy for solving problems, using a \textit{heuristic} $h(\state)$ to influence which state to \textit{expand}.
            In this work, we use two heuristic search algorithms implemented in RPID: \astar{}~\cite{DBLP:journals/tssc/HartNR68} and Complete Anytime Beam Search (CABS)~\cite{DBLP:conf/aaai/Zhang98}.
            In RPID, both \astar{} and CABS use the dual bound as $h(\state)$ and the cost to a node $g(\state)$ to guide the search, expanding the node with the minimum $f(\state) = g(\state) + h(\state)$.
            However, as opposed to \astar, CABS stores at most $b$ states with minimum $f(\state)$ in each \textit{layer} (storing at most $\mathcal{O}(n \times b)$ states at a time, where $n$ is the branching factor), increasing $b$ until termination.
            CABS has shown superior memory use and ability to prove optimality/infeasibility~\cite{KUROIWA2026104506} while also providing intermediate solutions.

            Pseudocode for heuristic search in DIDP can be seen in Algorithm~\ref{alg:heuristic_search_didp}; it starts with the target state, (heuristically) selecting a state to expand, and \textit{generating} its successors while avoiding dominated/redundant states, iterating this process until a termination condition is met.
            \begin{algorithm}
                \caption{Simplified Heuristic Search in RPID}
                \label{alg:heuristic_search_didp}
                \begin{algorithmic}
                    \State $\mathcal{O} \gets \{\targetstate\},\ \mathcal{E} \gets \emptyset,\ Primal \gets \infty$
                    \While{$\neg\mathtt{ShouldTerminate}()$}
                        \State $\state \gets \argmin_{\state \in \mathcal{O}} f(\state),\ \mathcal{O} \gets \mathcal{O} \setminus \{\state\}$%
                        \If{$\isbasestate{}$}
                            \State $Primal \gets \min\{Primal, g(\state) + \basecost\}$
                            \State \textbf{continue}
                        \EndIf
                        \For{$(w_\tau(\state), \state_{\predicate{\tau}}) \in \generatesuccessors$}
                            \State $h(\state_{\predicate{\tau}}) \gets \dualboundfuncinp{\state_{\predicate{\tau}}}$
                            \If{$f(\state_{\predicate{\tau}}) \leq Primal \land\ \nexists \state' \in \mathcal{E} $ s.t. 
                            \Statex \hspace{\algorithmicindent}\hspace{\algorithmicindent}$\dominates{\state'}{\state_{\predicate{\tau}}}\ \land\ g(\state') \leq g(\state_{\predicate{\tau}})$}
                                \State $\mathcal{O} \gets \mathtt{Insert}(\mathcal{O}, \state_{\predicate{\tau}}),\ \mathcal{E} \gets \mathcal{E} \cup \{\state_{\predicate{\tau}}\}$
                            \EndIf
                        \EndFor
                    \EndWhile
                \end{algorithmic}
            \end{algorithm}%
    \subsection{Constraint Programming}
        Constraint Programming (CP) is a paradigm for solving combinatorial problems in the form of a \textit{constraint satisfaction problem (CSP)},  represented by the tuple \((\mathcal{X}, \mathcal{C}, \mathcal{D})\), where \(\mathcal{X}\) is the set of \textit{variables}, \(\mathcal{C}\) is the set of \textit{constraints} specifying the relations between \textit{variables}, and \(\mathcal{D}\) is the \textit{domain} which maps $x \in \mathcal{X}$ to its possible values $\mathcal{D}(x) \subseteq \mathcal{Z}$.
        The goal is to find a \textit{solution} $\mathcal{I}$: a mapping of each variable $x \in \mathcal{X}$ to a value $v_x \in \mathcal{D}(x)$ which satisfies all constraints.

        A CSP can be transformed into a \textit{constraint optimisation problem (COP)} by adding an objective function which maps a solution $\mathcal{I}$ to a value: $o\colon \mathcal{I} \mapsto \mathcal{Z}$.
        The goal is then to find a solution that minimises (or maximises) this value.

        One strength of CP solvers is the use of inference techniques, which prune values from the domains of variables that cannot be part of a feasible solution.
        Concretely, constraints are represented in the CP solver by one or more \textit{propagators}.
        Thus, a propagator can be seen as a function $p\colon \mathcal{D} \mapsto \mathcal{D}'$ such that $\forall x \in \mathcal{X} : \mathcal{D}(x) \supseteq \mathcal{D}'(x)$, where a propagator is at \textit{fixed-point} if $p(\mathcal{D}) = \mathcal{D}$.
        We define \lb{x} (\ub{x}) as the minimum (maximum) value that can be assigned to $x \in \mathcal{X}$ in $\mathcal{D}$.
        Importantly, propagators are applied independently, but multiple propagators can be used together to express complex relationships between variables.
        
        In this work, we use two constraints and corresponding propagation algorithms, which are implemented in CP solvers such as \Pumpkin~\cite{flippo_et_al:LIPIcs.CP.2024.11}, Huub~\cite{dekker_et_al:LIPIcs.CP.2025.42}, and CP-SAT~\cite{cpsatlp}:
        
        The \disjunctive{} constraint~\cite{CARLIER198242}, states that a set of jobs should not overlap.
        Each job has a variable starting time and a (variable) duration.
        The most prevalent propagation algorithm is edge-finding~\cite{DBLP:conf/cpaior/Vilim04}, which reasons over whether scheduling a job $i$ as early (late) as possible is not possible due to another set of jobs $\Omega$ to infer that $i$ should be processed after (before) $\Omega$.

        The \cumulative{} constraint~\cite{DBLP:conf/jfplc/AggounB92}, states that the cumulative resource usage of a set of tasks should not exceed the capacity at any time point.
        Each task has a variable starting time, a duration, and resource usage.
        The most common propagation algorithm is time-table filtering~\cite{DBLP:conf/cp/BeldiceanuC02}, which reasons about when tasks are \textit{guaranteed} to consume resources to infer that another task \textit{cannot} be processed at these times.

\section{Related Work}
    Combining constraint propagation with dynamic programming was discussed by~\citeauthorandyear{DBLP:journals/constraints/HookerH18}, and while there has been work in this area, a generic approach for incorporating constraint propagation with heuristic search in a model-based way has remained  elusive.

    \citeauthorandyear{DBLP:journals/eor/FontaineDS23} apply propagation after each solution to tighten time windows when solving TSPTW using a variation of \astar~\cite{DBLP:conf/ausai/VadlamudiGAC12}. 
    Additionally, pruning techniques for DP approaches play a key role in reducing the search space for optimal decision tree problems~\cite{brița2025optimal}.
    While these approaches apply propagation within DP, their approach, unlike ours, relies on problem-specific propagation in a non-model-based approach.

    Generic propagation in a DP approach has been applied by~\citeauthorandyear{DBLP:journals/scheduling/TanakaF12}, making use of the \disjunctive{} for the \wttw{} problem.
    Their approach uses the Lagrangian relaxation to generate tight bounds based on a network representation, pruning states by propagating the \disjunctive{} constraint.
    However, as opposed to our work, they do not use a model-based approach, limiting the transferability of their techniques.

    From a CP perspective, the work by~\citeauthorandyear{DBLP:conf/cpaior/ChuBS10} proposes automatic identification and exploitation of subproblem dominance/equivalence, which can be used to prune states in a CP solver.
    In a similar vein,~\citeauthorandyear{DBLP:conf/cp/Smith05} proposes to create a representation of a key based on the (reduced) CP domains for caching.
    In our work, rather than adjusting the CP solver by adding dominance/equivalence constraints, we take the natural definition of dominance/equivalence in a DP model and incorporate propagation into this framework rather than the other way around, which means that we are able to more easily integrate with heuristic search using dual bounds.
    
    Furthermore, the work of \citeauthorandyear{DBLP:conf/cp/Perron99} transfers heuristic search to a CP solver.
    However, they do not leverage the dominance/duplication detection of the DP model.

    Lastly, \citeauthorandyear{DBLP:conf/ijcai/FoxSB89} proposed a model for problem solving by combining constraint satisfaction and heuristic search.
    However, the main difference between our work and theirs is that we perform constraint propagation for search in a generic model-based framework.

\section{DIDP with Constraint Propagation}
    We describe our framework for incorporating constraint propagation, and, for three problems, we describe the DP model and the incorporation of constraint propagation.

    \subsection{Framework}
        There are many possible architectures for combining solvers.
        Part of this space has been explored by the development of hybridisations of CP with fields such as Linear Programming~\cite{DBLP:journals/anor/BeckR03, DBLP:journals/scheduling/LaborieR16}, Satisfiability (SAT)~\cite{DBLP:journals/constraints/OhrimenkoSC09, DBLP:conf/cp/FeydyS09}, and Mixed-Integer Programming~\cite{DBLP:journals/constraints/Hooker05}.
        Additionally, there is extensive work on hybridisation architectures in the context of SAT Modulo Theories~\cite{DBLP:journals/jacm/NieuwenhuisOT06}.
        In our initial effort to develop a generic hybridisation of DP and CP, we chose to develop the simplest interface that would allow a DP solver to exploit the main technology of CP: constraint propagation.
        
        \begin{figure}[b]
            \centering
            \includegraphics[scale=1.0]{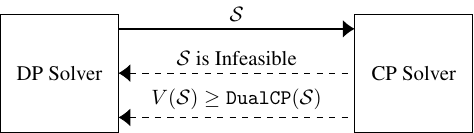}
            \caption{An overview of the interactions in our framework.}
            \label{fig:overview_framework}
        \end{figure}
        To this end, we utilise a generic CP solver implementing constraint propagation to provide information to the DP solver.
        An overview of the interactions between these two components can be seen in Figure~\ref{fig:overview_framework}.
        At its core, the CP solver provides two pieces of information: 1) when a state \state{} is infeasible, i.e., dynamically adding state/transition preconditions, and 2) a dual bound \dualboundcpfuncinp{\state}.
        It should be noted that this information can be determined by \emph{any} approach (e.g., decision diagram solvers or linear programs).
        
        To improve usability, we \textit{only} modify the RPID model and \textit{not} the search, by replacing \generatesuccessors{} in Algorithm~\ref{alg:heuristic_search_didp} with \generatesuccessorswithpropagation{} (Algorithm~\ref{alg:gen_succ_with_propagation}), and \dualboundfuncinp{\state_{\predicate{\tau}}} with $\max\{\dualboundfuncinp{\state_{\predicate{\tau}}}, \dualboundcpfuncinp{\state_{\predicate{\tau}}, \mathcal{D}'}\}$.
        Algorithm~\ref{alg:gen_succ_with_propagation} first creates the CP model, which is propagated to obtain $\mathcal{D}'$ and checked for infeasibility using \isstateinfeasible{} and \dualboundcpfuncinp{\state, \mathcal{D}'}; one benefit of using CP is that $\mathcal{D}'$ can be used to strengthen the dual bound.
        Next, successor infeasibility is checked in \issuccessorinfeasible; since $\mathcal{D}'$ is valid for any successor $\state_{\predicate{\tau}}$, $\mathcal{D}'$ is used in \issuccessorinfeasible{} to avoid propagation.
        In practice, we only initialise $\mathcal{X}$ and $\mathcal{C}$ for the target state, and adjust $\mathcal{D}$ to represent every state.
        \begin{algorithm}
            \caption{\generatesuccessorswithpropagation}
            \label{alg:gen_succ_with_propagation}
            \begin{algorithmic}
                \State $(\mathcal{X}, \mathcal{C}, \mathcal{D}) \gets \createcpmodel$
                \State $\mathcal{D}' \gets \propagate$
                \If{$\isstateinfeasible \lor\ g(\state) + \dualboundcpfuncinp{\state, \mathcal{D}'} \geq Primal$}
                    \State \Return $\emptyset$
                \EndIf
                \State $Successors \gets \emptyset$
                \For{$(w_\tau(\state), \state_{\predicate{\tau}}) \in \generatesuccessors$}
                    \If{$\neg \issuccessorinfeasible$}
                        \State $Successors \gets Successors \cup \{(w_\tau(\state), \state_{\predicate{\tau}})\}$
                    \EndIf
                \EndFor
                \State \Return $Successors$
            \end{algorithmic}
        \end{algorithm}%
    
    \subsection{$\bm{\wttw}$}
        The single-machine scheduling problem \wttw{} is formulated as follows: given are $n$ jobs $\jobs = \{j_0, \cdots, j_{n - 1}\}$ which need to be processed on a single machine without overlapping.
        Each job $i \in \jobs$ has a duration $p_i \in \mathcal{N^+}$, a release time $r_i \in \mathcal{N}$, a deadline $d_i \in \mathcal{N}$, a weight $w_i \in \mathcal{N}$, and a latest possible finish time $\delta_i \in \mathcal{N}$.
        
        The goal is to assign a start time $s_i$ to each job such that $r_i \leq s_i < s_i + p_i \leq \delta_i$, minimising the weighted tardiness $\sum_{i \in \jobs} w_i \tardiness$, where $\tardiness = \max\{0, s_i + p_i - d_i\}$.

        \subsubsection{DP Model}
            The DP model (adapted from~\citeauthorandyear{ABDULRAZAQ1990235}) schedules one job at a time.
            We track the unscheduled jobs \unscheduled{} and current time \current{} in $\statewttw = (\unscheduled, \current)$.
            Additionally, we define the next available time after scheduling $i$ as $\current'(\current, i) = \max\{\current,\ r_i\} + p_i$, the set of unscheduled jobs for which the time-window is not violated when scheduled $B = \{i \in \unscheduled : \current'(\current, i) \leq \delta_i\}$, and the successor state $\statewttw_{\predicate{i}} = (\unscheduled \setminus \{i\}, \current'(\current, i))$.
            
            The Bellman equation is then defined as:
            \begin{flalign}
                &\valuef{\unscheduled, \current} = 
                \makebox[0pt][l]{$\begin{cases}
                  0 & \text{if } \unscheduled = \emptyset\\
                  \infty & \text{if } \unscheduled \neq B\\
                  \min_{i \in \unscheduled} w_i \times \tardiness + \valuef{\statewttw_{\predicate{i}}}
                  & \text{else}
                \end{cases}$}&\label{eq:wttw_recurrence}\\
                & \valuef{\unscheduled, \current^a} \leq \valuef{\unscheduled, \current^b} & \text{if }\current^a \leq \current^b\label{eq:wttw_dominance}&\\
                & \makebox[0pt][l]{$\valuef{\unscheduled, \current} \geq \sum\nolimits_{i \in \unscheduled} w_i \times \max\{0, \max\{r_i, \current\} + p_i - d_i\}$}&\label{eq:wttw_dual_bound}
            \end{flalign}
            Equation~\eqref{eq:wttw_recurrence} states that a state is infeasible if there are jobs that cannot be scheduled due to time windows.
            Inequality~\eqref{eq:wttw_dominance} states that state $a$ dominates state $b$ if state $a$ schedules the same jobs as state $b$ but does so in less time.
            Inequality~\eqref{eq:wttw_dual_bound} describes a dual bound as the sum of tardinesses if all unscheduled jobs were started at $\max\{r_i, \current\}$.

        \subsubsection{Constraint Propagation}
            The variables $\mathcal{X}$ and domains $\mathcal{D}$ created in \createcpmodel{} are described in Definition~\eqref{eq:wttw_cp_variables}; for each unscheduled job $i$, a CP variable for the start time is created, which starts after $r_i$ \textit{and} all scheduled jobs.
            
            The CP constraints $\mathcal{C}$ are then defined by
            Constraint~\eqref{eq:wttw_disjunctive}, which uses the \disjunctive{} to ensure that no jobs overlap.
            \begin{flalign}
                & \disjunctive([s_i\ |\ i \in \unscheduled], [p_i\ |\ i \in \unscheduled]) & \label{eq:wttw_disjunctive}\\
                & s_i \in [\max\{r_i,\ t\},\ \delta_i - p_i] & \forall i \in \unscheduled & \label{eq:wttw_cp_variables}
            \end{flalign}
            
            After \propagate, we use the reduced domains $\mathcal{D}'$ in \dualboundcpfuncinp{\state, \mathcal{D}'} as shown in Inequality~\eqref{eq:wttw_cp_dual_bound}, which states a dual bound as the sum of tardinesses of all unscheduled jobs if started at $\lb{s_i}$.
            Additionally, Equation~\eqref{eq:wttw_feasibility} describes \issuccessorinfeasible{}, where a successor is infeasible if $\tau$ cannot be scheduled at $\current'(\current, \tau)$.
            \begin{flalign}
                & \makebox[0pt][l]{$\valuef{\unscheduled, \current} \geq\ \sum\nolimits_{i \in \unscheduled} w_i \times \max\{0, \lb{s_i} + p_i - d_i\}$}&\label{eq:wttw_cp_dual_bound}\\
                & \valuef{\state_{\predicate{\tau}}} = \infty & \text{if }\current'(t, \tau) \notin \mathcal{D}'(s_\tau) &\label{eq:wttw_feasibility}
            \end{flalign}
        \subsection{RCPSP}
        The Resource-Constrained Project Scheduling Problem can be formulated as follows: given are $n$ tasks $\tasks = \{t_0, \cdots, \allowbreak t_{n - 1}\}$ and a set of resources \resources{}, where each resource $r \in \resources$ has a capacity $\capacity_r \in \mathcal{N}^+$.
        Each task has a duration $p_i \in \mathcal{N^+}$, and a resource usage $u_{ir} \in \mathcal{N}$ per resource $r \in \resources$.

        The goal is to assign a start time $s_i \in \mathcal{N}$ to each task such that 1) the cumulative resource usage never exceeds \textit{any} resource capacity, and 2) a set of precedence constraints is respected, as shown in Inequalities~\eqref{eq:cumulative}-\eqref{eq:precedence} (where $\horizon = \sum_{i \in \tasks} p_i$).
        We define the predecessors of task $i$ as $\predecessors_i$.
        The objective is to minimise the makespan: $\max_{i \in \tasks} s_i + p_i$.
        \begin{flalign}
            & \sum_{i \in \tasks\ :\ s_i \leq t\ \land\ s_i + p_i \geq t} u_{ir} \leq \capacity_r & \forall t \in [0, \horizon],\ r \in \resources\ \label{eq:cumulative}\\
            & s_i + p_i \leq s_j & \forall (i,j) \in \precedences\label{eq:precedence}
        \end{flalign}

        \subsubsection{DP Model}
            We introduce a DP model based on scheduling one task at a time.
            We keep track of the partial schedule \partialschedule{} mapping each scheduled task $j$ to its scheduled start time $\partialschedule_j$ (and each unscheduled task to $\bot$), and the current time \current{} in $\statercpsp = (\partialschedule, \current)$.
            We define the set of scheduled tasks $\scheduled = \{i\ |\ i \in \tasks : \partialschedule_i \neq \bot\}$ (and $\unscheduled = \tasks \setminus \scheduled$).

            Equation~\eqref{eq:rcpsp_time_point} defines $\current'(\current, \tau)$ as the earliest time where 1) there is no resource conflict when scheduling $\tau$ and 2) all of the predecessors of $\tau$ have finished.
            We then define the successor state as $\statercpsp_{\predicate{\tau}} = ((\partialschedule \setminus \{(\tau, \bot)\}) \cup \{(\tau, \current'(\current, \tau)\}, \current'(\current, \tau))$;  we also define $\makespan{\partialschedule, \current} = \max\{\max_{i \in \scheduled}\{\partialschedule_i + p_i\},  \max_{i \in \unscheduled}\{t + p_i\}\}$, and $\weight = (\makespan{\statercpsp_{\predicate{\tau}}, \current'(\current, \tau)} - \makespan{\statercpsp, \current})$.%
            \begin{equation}
                \label{eq:rcpsp_time_point}
                \current'(\current, \tau) =
                \begin{cases}
                    \min h \in [\current, \horizon] \text{ s.t. } \nexists r \in \resources\ :\\
                    u_{\tau r} + \sum\nolimits_{i \in \scheduled\ :\ s_i \leq h\ \land\ s_i + p_i > h} u_{ir} > \capacity_r\\
                    \land\ \predecessors_{\tau} \subseteq \{j\ |\ j \in \scheduled \land s_j + p_j \leq h\}
                \end{cases}              
            \end{equation}
            The Bellman equation is then defined as:
            \begin{align}
                 & \valuef{\partialschedule, \current} = 
                \begin{cases}
                  0 & \text{if } \scheduled = \tasks\\
                  \min\limits_{i \in \unscheduled : \predecessors_i \subseteq \scheduled} \weight + \valuef{\statercpsp_{\predicate{\tau}}} & \text{if } \scheduled \neq \tasks
                \end{cases}
            \end{align}
            \begin{align}
                & \valuef{\state^a} \leq \valuef{\state^b} \quad \text{if }
                    \begin{cases}
                        \scheduled^a = \scheduled^b\ \land\ t^a \leq t^b\ \land\\
                        \forall i \in \scheduled : \max(\partialschedule^a_j, \partialschedule^b_j) \\ + p_j > \current
                        \rightarrow \partialschedule^a_i \leq \partialschedule^b_i
                    \end{cases}\label{eq:dominance}
            \end{align}
            \begin{flalign}
                & \valuef{\state_{\predicate{\tau}}} \leq \valuef{\state_{\predicate{\tau'}}} & \text{if } \current'(\current, \tau) + p_\tau \leq  \current'(\current, \tau')&\label{eq:left-shift}
            \end{flalign}
            \begin{flalign}
                & \valuef{\partialschedule, \current} \geq \mathtt{Length}(\text{CriticalPath}(\precedences, \unscheduled))\label{eq:rcpsp_critical_path_bound}&\\
                & \valuef{\partialschedule, \current} \geq \max\limits_{r \in \resources} \left\lceil\frac{\sum_{i \in \unscheduled} u_{ir} \times p_i} {\capacity_r}\right\rceil&\label{eq:rcpsp_energy_bound}
            \end{flalign}
            Inequality~\eqref{eq:dominance} describes dominance between states $a$ and $b$; if they have scheduled the same tasks but $a$ has scheduled all tasks sooner than $b$, then $a$ dominates $b$.
            Similarly, Inequality~\eqref{eq:left-shift} describes a transition dominance using the left-shift rule~\cite{4bd0a2b3-d0f9-30de-9550-36586a44cbdc}; if there are two transitions $\tau, \tau' \in \mathtt{T}(\state)$, then $\state_{\predicate{\tau}}$ dominates $\state_{\predicate{\tau'}}$ if we can complete task $\tau$ \textit{before} starting $\tau'$.
            Inequality~\eqref{eq:rcpsp_critical_path_bound} states that \valuef{\state} is larger than the length of the longest path through the precedence graph (i.e., the critical path).
            Inequality~\eqref{eq:rcpsp_energy_bound} states an energy-based dual-bound.

        \subsubsection{Constraint Propagation}
            The variables $\mathcal{X}$ and domains $\mathcal{D}$ created in \createcpmodel{} are described in Definitions~\eqref{eq:rcpsp_scheduled}-\eqref{eq:rcpsp_objective_var}; for each task $i$, a CP variable $s_i$ for the start time is created, and a CP variable $o$ for the objective is created.
            
            The CP constraints $\mathcal{C}$ are defined as:
            Constraint~\eqref{eq:rcpsp_cumulative} specifies the \cumulative{} which ensures the resource capacity is not exceeded (where $\starttimes{} = [s_i\ |\ i \in \unscheduled]$, $\processingtimes{} = [p_i\ |\ i \in \unscheduled]$, and $\resourceusages_r = [u_{ir}\ |\ i \in \unscheduled]$), Constraint~\eqref{eq:rcpsp_precedences} ensures that precedences hold, and Constraint~\eqref{eq:rcpsp_objective} defines the makespan.
            \begin{flalign}
                &\cumulative(\starttimes, \processingtimes, \resourceusages_r, \capacity_r) & \forall r \in \resources\label{eq:rcpsp_cumulative}\\
                & s_i + p_i \leq s_j & \forall (i, j) \in \precedences\label{eq:rcpsp_precedences}\\
                & s_i + p_i \leq o \leq Primal & \forall i \in \unscheduled \label{eq:rcpsp_objective}\\
                & s_j = \partialschedule_j & \forall j \in \scheduled\label{eq:rcpsp_scheduled}\\
                & s_i \in [\current, \horizon - p_i] & \forall i \in \unscheduled\label{eq:rcpsp_unscheduled}\\
                & o \in [0, \horizon] \label{eq:rcpsp_objective_var}
            \end{flalign}
            
            After \propagate, we use the bounds $\mathcal{D}'$ in \dualboundcpfuncinp{\state, \mathcal{D}'} as described in 1) 
            Inequality~\eqref{eq:rcpsp_edge_finding_bound}, which specifies a lower-bound on the makespan based on the earliest completion time~\cite{DBLP:conf/cp/Vilim09}, and 2) Inequality~\eqref{eq:rcpsp_simple_cp_bound}, which states that the makespan of a state is at least as large as the maximum latest finish time of the unscheduled tasks.
            Additionally, Equation~\eqref{eq:rcpsp_infeasibility} describes \issuccessorinfeasible{}, where a successor is not feasible if task $\tau$ cannot be scheduled at time $\current'(\current, \tau)$.
            \begin{flalign}
                & \makebox[0pt][l]{$\valuef{\statercpsp} \geq \max\limits_{r \in \resources}\max\limits_{\Omega \subseteq \tasks}\{\left\lceil \frac{\capacity_r \min\limits_{i \in \Omega} \lb{s_i} + \sum\limits_{i \in \Omega} (u_{ir} \times p_i)}{\capacity_r} \right\rceil\}$}&\label{eq:rcpsp_edge_finding_bound}\\
                & \makebox[0pt][l]{$\valuef{\partialschedule, \current} \geq \max\nolimits_{i \in \unscheduled}\{\lb{s_i} + p_i\}$} \label{eq:rcpsp_simple_cp_bound}\\
                & \valuef{\statercpsp_{\predicate{\tau}}} = \infty & \text{if }\current'(\current, \tau) \notin \mathcal{D}'(s_\tau)\label{eq:rcpsp_infeasibility}
            \end{flalign}
    \subsection{TSPTW}
        The Travelling Salesperson Problem with Time Windows can be formulated as follows: given are $n$ locations $\locations = \{l_0, \cdots, l_{n - 1}\}$ which need to be visited, with a travel time $\distance{i}{j} \in \mathcal{N}$ between locations $i$ and $j$.
        Each location $i$ needs to be visited within $[r_i, \delta_i]$; if the salesperson arrives at location $i$ before $r_i$ then they wait until time $r_i$.
        We define $\distance{i}{j}^*$ as the shortest travel time between location $i$ and $j$.

        The goal is to determine the order in which to visit the locations, starting and ending at the depot $l_0$ while visiting each location besides the depot exactly once, such that the total travel time is minimised (excluding waiting times).

        \subsubsection{DP Model}
            The DP model~\cite{KUROIWA2026104506} is based on visiting one location at a time.
            We track the unvisited locations \unvisited, current location \currentlocation, and current time \current{} in $\statetsptw = (\unscheduled, \currentlocation, \current)$.
            Additionally, we define $\current'(i, j) = \max\{\current + \distance{i}{j}, r_j\}$, $\state_{\predicate{j}} = (\unvisited \setminus \{j\}, j, \current'(\currentlocation, j))$, $\violated = \{j\ |\ j \in \unvisited\ :\ \current + \distance{i}{j}^{*} > \delta_j\}$, $\minto{l} = \min\nolimits_{i \in \locations : i \neq j} c_{il}$, $\minfrom{l} = \min\nolimits_{i \in \locations : i \neq j} c_{li}$, and $\reachable_i = \{j \in \unvisited\ |\ \current + \distance{i}{j} \leq \delta_j\}$.

            The Bellman equation is then defined as:
            \begin{flalign}
                & \valuef{\statetsptw} = 
                \makebox[0pt][l]{$\begin{cases}
                    \infty & \text{if }\violated \neq \emptyset \\
                    \distance{l}{0} & \text{if } \unvisited = \emptyset\\
                    \min_{j \in \reachable_\currentlocation} \distance{\currentlocation}{j} + \valuef{\state_{\predicate{j}}} & \text{else}
                \end{cases}$}&\\
                & \valuef{\unscheduled, \currentlocation, \current_a} \leq \valuef{\unscheduled, \currentlocation, \current_b} & \text{if }\current_a \leq \current_b&\label{eq:tsptw_dominance}\\
                & \valuef{\unscheduled, \currentlocation, \current} \geq                
                    \makebox[0pt][l]{$\begin{cases}
                            \minto{l_0} + \sum_{i \in \unvisited} \minto{i}\\
                            \minfrom{\currentlocation} + \sum_{i \in \unvisited} \minfrom{i}
                         \end{cases}$}&\label{eq:tsptw_dual_bound}
            \end{flalign}
            Inequality~\eqref{eq:tsptw_dominance} states that state $a$ dominates state $b$ if 1) they have visited the same locations, 2) are at the same current location, and 3) the current time of state $a$ is earlier than the current time of state $b$.
            Inequality~\eqref{eq:tsptw_dual_bound} uses the minimum travel time \textit{to} an unvisited location $i$ from another location and the minimum travel time \textit{from} an unvisited location $i$ to another location to calculate a dual bound on \statetsptw.
            
        \subsubsection{Constraint Propagation}
            The variables $\mathcal{X}$ and domains $\mathcal{D}$ created in \createcpmodel{} are described in Definitions~\eqref{eq:tsptw_start}-\eqref{eq:tsptw_objective_var}; for each unvisited location $i$, a CP variable $s_i$ for the arrival time and time to next location $p_i$ is created, where the bounds on $p_i$ can be tightened if $\exists j \in \unscheduled \setminus \{i\} : \lb{s_j} \geq \ub{s_i}$, signifying that the depot cannot be visited from $i$. 
            Additionally, a CP variable $o$ for the objective is created.
            
            The constraints $\mathcal{C}$, a relaxation of TSPTW, are defined as 1) Constraint~\eqref{eq:tsptw_disjunctive} stating that no locations can be visited simultaneously, and 2) Constraint~\eqref{eq:tsptw_objective} defining the objective.
            \begin{flalign}
                & \makebox[0pt][l]{$\disjunctive([s_i\ |\ i \in \unscheduled \cup \{\currentlocation\}], [p_i\ |\ i \in \unscheduled \cup \{\currentlocation\}])$}&\label{eq:tsptw_disjunctive}\\
                & o = \sum\nolimits_{i \in \locations} p_i \leq Primal & \label{eq:tsptw_objective}\\
                & s_i \in [\max\{\current, r_i\}, \delta_i] & \forall i \in \unscheduled \cup \{\currentlocation\}\label{eq:tsptw_start}
            \end{flalign}
            \begin{flalign}
                & p_i \in \{\distance{i}{j}|j \in (\unscheduled \cup \{l_0\}) \setminus \{i\} \land \distance{i}{j} \neq \infty\} & \forall i \in \unscheduled \cup \{\currentlocation\}\label{eq:tsptw_duration}\\
                & o \in [0, \max\nolimits_{i \in \unvisited} \delta_i + \distance{i}{0}]\label{eq:tsptw_objective_var}
            \end{flalign}

            After \propagate, we use the bounds $\mathcal{D}'$ in \dualboundcpfuncinp{\state, \mathcal{D}'} as described in Inequality~\eqref{eq:tsptw_cp_dual_bound}, which states that a dual bound is the sum of the lower bounds of the travel times inferred by constraint propagation.
            Additionally, Equation~\eqref{eq:tsptw_feasibility} describes \issuccessorinfeasible{}, where visiting $\tau$ next is infeasible if its arrival time or the travel time from the previous location is not possible.%
            \begin{flalign}
                & \makebox[0pt][l]{$\valuef{\unscheduled, \currentlocation, \current} \geq \sum\nolimits_{i \in \unscheduled \cup{\currentlocation}} \lb{p_i}$} &\label{eq:tsptw_cp_dual_bound}\\
                & \valuef{\state_{\predicate{\tau}}} = \infty & \text{if } \current'(\currentlocation, \tau) \notin \mathcal{D}'(s_\tau) \lor \distance{\currentlocation}{\tau} \notin \mathcal{D}'(p_\tau) &\label{eq:tsptw_feasibility} 
            \end{flalign}
\section{Experimentation}
    Our aim is to empirically show the impact of constraint propagation on the number of state expansions and runtime when integrated with a DP solver.
    For all three problems, we show that constraint propagation reduces the search space.
    For \wttw{} and TSPTW, we show that it is especially effective for tightly constrained instances.
    
    \subsection{Experimentation Setup}
        The experiments are run single threaded on an Intel Xeon Gold 6248R 24C 3.0GHz processor~\cite{DHPC2024} 
        with a limit of 30 minutes and 16GB of memory.
        We use the following benchmarks:
        \begin{itemize}
            \item $\bm{\wttw}$ - 
                We generate 900 instances according to~\citeauthorandyear{DBLP:journals/scheduling/DavariDLN16}.
                The instances contain 50 jobs with duration $p_i \in [1, 10]$, release date $r_i \in [0, \tau P]$ (where $P$ is the sum of durations), deadline $d_i \in [r_i + p_i, r_i + p_i + \rho P]$, latest finish time $\delta_i \in [d_i, d_i + \phi P]$, and weight $w_i \in [1, 10]$.
                We generate 10 instances for each combination of $\tau \in \{0, 0.2, 0.4, 0.6, 0.8, 1\}$, $\rho \in \{0.05, 0.25, 0.5\}$, $\phi \in \{0.9, 1.05, 1.2, 1.35, 1.5\}$.
            \item \textbf{RCPSP} - 
                We make use of the 480 J90 instances from PSPLIB~\cite{KOLISCH1997205}.
            \item \textbf{TSPTW} -
                We consider instances used by \citeauthorandyear{DBLP:journals/asc/Lopez-IbanezBOT13}, excluding too easy/difficult sets or sets with fractional distances; resulting in 130 instances by~\citeauthorandyear{DBLP:journals/ior/GendreauHLS98} and 50 instances by~\citeauthorandyear{Ascheuer1996}.
        \end{itemize}

        We use \Pumpkin~\cite{flippo_et_al:LIPIcs.CP.2024.11} for propagation, and CABS and \astar{} of RPID 0.3.1 for the DP-based approaches.
        The code and models are in the supplements.
        We evaluate the following approaches (only considering generic model-based approaches, not problem-specific ones, to gauge the benefit of propagation in \textit{general-purpose} DP solvers):
        \begin{itemize}
            \item 
                \textbf{\astar/CABS} - The RPID model using \astar{} or CABS.
            \item 
                \textbf{\astar/CABS+CP} - The RPID model using \astar{} or CABS \textit{with} constraint propagation; the propagators are executed once per state.
                We also considered fixed-point propagation, but the computational cost was too high to be able to compete with other methods.
            \item 
                \textbf{ORT} - Uses MiniZinc 2.9.2 with OR-Tools CP-SAT 9.12 as solver (using free search, causing the solver to make use of a portfolio approach); included in the comparison as a reference state-of-the-art constraint-based solver.
                The CP model descriptions are in Appendix~\potentialremove{\ref{sec:cp_models}}{B}.
        \end{itemize}

    \paragraph*{Experimentation Summary}
        Overall, constraint propagation significantly prunes the search space for all problems.
        For \wttw{}, RCPSP, and tightly constrained TSPTW problems, our approach solves more instances than the DP approach while using considerably fewer states.
        The runtime performance indicates that further work on reducing propagation time is warranted.

    \subsection{$\bm{\wttw}$}
\begin{figure}[bt]
            \begin{subfigure}[htbp]{\columnwidth}
                \centering
                \includegraphics[scale=1.0]{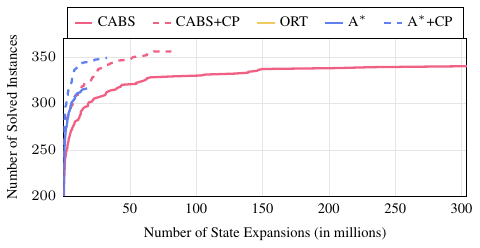}
                \caption{\wttw{} instances solved compared to state expansions. Our approach is able to solve the most instances per expansion.}
                \label{fig:wttw_expanded}
            \end{subfigure}
            \begin{subfigure}[htbp]{\columnwidth}
                \centering
                \includegraphics[scale=1.0]{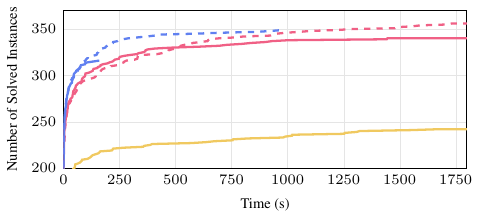}
                \caption{\wttw{} instances solved compared to time. Our approach solves the most instances.}
                \label{fig:wttw_time}
            \end{subfigure}
            \caption{
                Number of \wttw{} instances solved compared to the number of state expansions and time.
            }
            \label{fig:wttw_overview}
        \end{figure}

        Starting with \wttw, Figure~\ref{fig:wttw_expanded} shows that our approach solves significantly more instances using fewer states than the baseline versions, exhibiting the effectiveness of constraint propagation.
        This effectiveness is confirmed by Appendix~\potentialremove{\ref{sec:wttw_opt_gap}}{A.1}, which shows that our approach guides the search to better solutions in fewer expansions.
        While \astar+CP initially solves more instances, it plateaus sooner than CABS+CP, ultimately solving fewer instances.
        
        Looking at the number of instances solved over time, Figure~\ref{fig:wttw_time} shows that the best performing approach is CABS+CP, which proves optimality for three more instances and infeasibility (at the target state) on thirteen more than CABS.
        Interestingly, CABS+CP only overtakes CABS in the number of instances solved after 500 seconds, showing the impact of propagation overhead.
        Moreover, the lack of instances solved by OR-Tools indicates that it is our combination of DP \textit{with propagation} that works well. 
      
        \subsubsection{Parameter Analysis}
\begin{figure}[tb]
                \centering
                \includegraphics[scale=1.0]{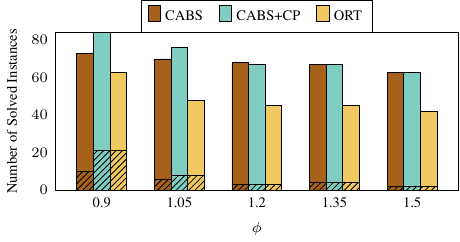}
                \caption{Number of \wttw{} instances solved over $\phi$. The marked parts are instances proven infeasible and the unmarked parts are instances proven optimal. Our approach solves the most instances when they are tightly constrained.}
                \label{fig:wttw_parameters}
            \end{figure}

            An overview of the number of instances solved across values of $\phi$ (where a lower value means that the instances have tighter latest finish times), can be seen in Figure~\ref{fig:wttw_parameters}.
            It shows that CABS+CP outperforms CABS on constrained instances where $\phi$ is small(er), indicating that the constraint propagation prunes well when instances are tightly constrained.
            Furthermore, the number of solved instances decreases as $\phi$ increases for all approaches, but the rate of decrease differs.
            Specifically, when $\phi \leq 1.05$, CABS+CP outperforms CABS in terms of instances proven infeasible \textit{and} optimal, but for instances where $\phi \geq 1.2$, CABS is equal to or outperforms CABS+CP.

    \subsection{RCPSP}
\begin{figure}[bthp]
            \begin{subfigure}[htbp]{\columnwidth}
                \centering
                \includegraphics[scale=1.0]{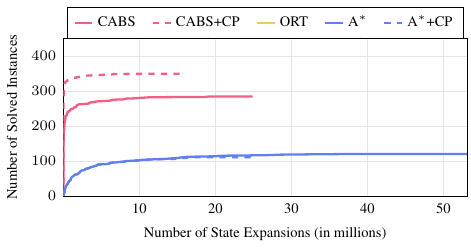}
                \caption{RCPSP instances solved compared to  state expansions. Our approach solves significantly more instances per state expansion.}
                \label{fig:rcpsp_expanded}
            \end{subfigure}
            \begin{subfigure}[htbp]{\columnwidth}
                \centering
                \includegraphics[scale=1.0]{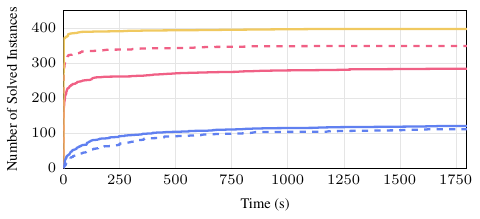}
                \caption{RCPSP instances solved compared to time; our approach outperforms the DP baseline, but the best-performing approach is CP.}
                \label{fig:rcpsp_time}
            \end{subfigure}
            \caption{
                Number of RCPSP instances solved compared to the number of state expansions and time.
            }
            \label{fig:rcpsp_overview}
        \end{figure}

        Considering RCPSP, Figure~\ref{fig:rcpsp_expanded} shows that constraint propagation increases the number of instances solved per state expansion considerably for CABS, exhibiting that constraint propagation is crucial when using a state-based approach for RCPSP.
        Appendix~\potentialremove{\ref{sec:rcpsp_opt_gap}}{A.2} indicates that constraint propagation provides the most benefit when proving optimality.
        
        Additionally, Figure~\ref{fig:rcpsp_time} shows that CABS+CP solves more instances per second than CABS alone.
        Since \astar{}(+CP) rarely finds primal bounds, which are key for pruning, it falls behind CABS(+CP).
        OR-Tools is the best-performing solver, possibly due to its depth-first backtracking search being well-suited for the problem.
        Nonetheless, \astar/CABS+CP provide better primal and/or dual bounds than OR-Tools on 70 instances (as shown in Figure~\potentialremove{\ref{fig:rcpsp_opt_gap_comparison}}{11} of Appendix~\potentialremove{\ref{sec:rcpsp_opt_gap}}{A.2}).

    \subsection{TSPTW}
    \label{sec:tsptw}
\begin{figure}[bthp]
        \begin{subfigure}[htbp]{\columnwidth}
                \centering
                \includegraphics[scale=1.0]{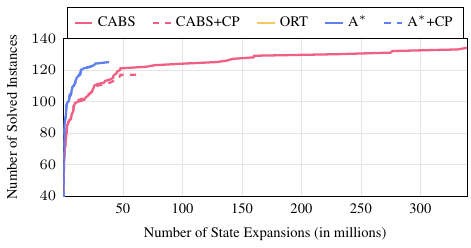}
                \caption{TSPTW instances solved compared to state expansions; the line for \astar+CP fully overlaps with \astar{}. Our approach expands slightly fewer states than the DP approach.}
                \label{fig:tsptw_expanded}
            \end{subfigure}
            \begin{subfigure}[htbp]{\columnwidth}
                \centering
                \includegraphics[scale=1.0]{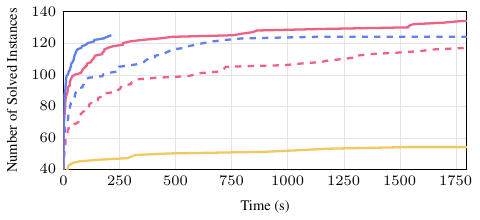}
                \caption{TSPTW instances solved compared to time. The DP approach solves the most instances.}
                \label{fig:tsptw_time}
            \end{subfigure}
            
            \caption{
                Number of TSPTW instances solved compared to the number of state expansions and time.
            }
            \label{fig:tsptw_overview}
        \end{figure}

        Turning to TSPTW, Figure~\ref{fig:tsptw_expanded} shows that constraint propagation \emph{decreases} the number of solved instances due to a lack of pruning.
        This observation is demonstrated in the number of solved instances per state expansion of both \astar{}/\astar+CP and CABS/CABS+CP overlapping, likely due to instances being too loosely constrained to make inferences.
        This observation is further corroborated by Appendix~\potentialremove{\ref{sec:tsptw_opt_gap}}{A.3}.
        
        This effect is also reflected by the number of instances solved over time in Figure~\ref{fig:tsptw_time}, which shows that adding constraint propagation results in an increase in solving time.
        However, the worst-performing approach is OR-Tools.

        \subsubsection{Parameter Analysis}
            To determine the impact of instance constrainedness, we investigate parameterised instances (not part of the previous set) introduced by~\citeauthorandyear{DBLP:journals/jair/RifkiS25} where $n = 31$.
            For these instances, a low $\alpha$ indicates that many locations need to be visited relative to the horizon, while a low $\beta$ means that time windows are tight.
            We consider two sets: 1) a general set of 135 instances where we select three instances per combination of $\alpha \in \{1.0, 1.5, 2, 2.5, 3\}$ and $\beta \in \{0.2, 0.3, 0.4, 0.5, 0.6, 0.7, 0.8, 0.9, 1\}$, and 2) a set of 540 instances where we select ten instances per combination of $\alpha \in \{1.0, 1.1, 1.2, 1.3, 1.4, 1.5\}$ and the same values of $\beta$.

\begin{figure}[tb]
                \centering
                \begin{subfigure}[htbp]{\columnwidth}
                    \hspace*{0.1cm}\includegraphics[scale=1.0]{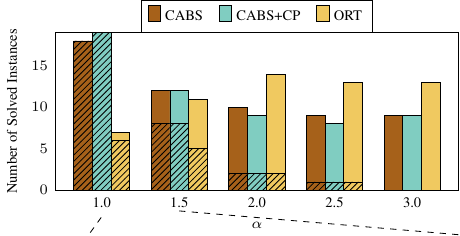}
                        \label{fig:tsptw_parameters}
                \end{subfigure}%

                \vspace*{-0.25cm}\begin{subfigure}[htbp]{\columnwidth}
                    \centering
                    \includegraphics[scale=1.0]{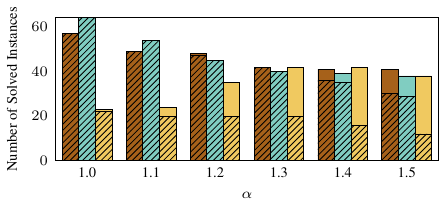}
                    \label{fig:tsptw_parameters_alpha}
                \end{subfigure}
                \caption{Number of solved TSPTW \citeauthorandyear{DBLP:journals/jair/RifkiS25} instances across $\alpha$. The marked parts are instances proven infeasible and the unmarked parts are instances proven optimal. The lower plot uses an extended set of instances to show $1.0 \leq \alpha \leq 1.5$.
                Our approach solves the most instances when they are tightly constrained.}
                \label{fig:tsptw_parameter_analysis}
            \end{figure}

\begin{figure}[htb]
            \centering
                \includegraphics[scale=1.0]{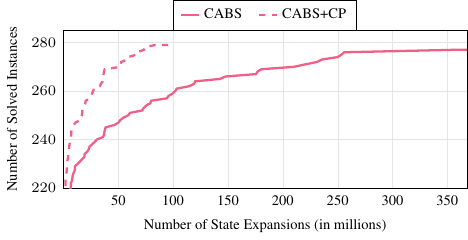}
                \caption{Number of solved TSPTW~\citeauthorandyear{DBLP:journals/jair/RifkiS25} instances ($1.0 \leq \alpha \leq 1.5$) over state expansions. Our approach solves the most instances per expansion.}
                \label{fig:tsptw_specific_expanded}
        \end{figure}

\begin{figure}[htb]
            \includegraphics[scale=1.0]{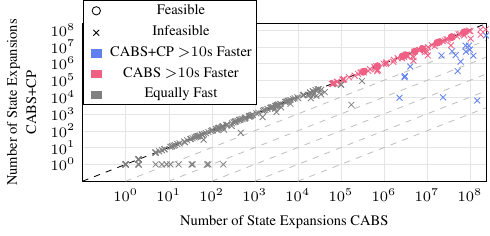}
            \caption{Comparison of state expansions on TSPTW~\citeauthorandyear{DBLP:journals/jair/RifkiS25} instances ($1.0 \leq \alpha \leq 1.5$). Our approach reduces the number of expansions by orders of magnitude.}
            \label{fig:tsptw_specific_comparison}
        \end{figure}

            Figure~\ref{fig:tsptw_parameter_analysis}, shows that for $\alpha < 1.2$, CABS+CP outperforms all other approaches.
            Examining Figure~\ref{fig:tsptw_specific_expanded}, shows that, when $1.0 \leq \alpha \leq 1.5$, the number of instances solved per state expansion is significantly increased compared to Figure~\ref{fig:tsptw_expanded}.
            This effect can be explained by CABS+CP reducing the number of states by orders of magnitude compared to CABS (Figure~\ref{fig:tsptw_specific_comparison}), where especially infeasible instances benefit from constraint propagation.
            Overall, the DP-based methods solve fewer instances as $\alpha$ increases, but the rate of decrease differs, causing CABS to outperform CABS+CP when $\alpha \geq 1.2$.
            Interestingly, OR-Tools proves optimality on the most instances, likely due to 1) lower memory usage, and 2) stronger propagation (using \constraint{Circuit}~\cite{DBLP:journals/ai/Lauriere78}).
        
\section{Conclusions and Future Work}
    We have presented a framework that integrates constraint propagation into a DP solver, and implemented it in the modelling of DIDP.
    Our approach combines the strengths of heuristic guidance and dominance/duplicate detection of DP with the inference techniques of CP.
    Our evaluation on three combinatorial optimisation problems shows that adding constraint propagation significantly increases the number of instances solved per state expansion, solving more instances than a DP solver alone for \wttw{}, RCPSP, and tightly constrained TSPTW instances.
    As for runtime, the benefit of propagation outweighs the cost for tightly constrained instances, but work into reducing propagation time could improve performance further. 
    In our in-depth analysis, we vary \wttw{} and TSPTW instance constrainedness, showing that propagation performs well when instances are highly constrained.
    Our work is a key step in understanding the value of constraint propagation in DP solvers, providing a model-based integration of DP and CP.

    Our interface is generic, allowing future work to investigate the integration of DP solvers with other techniques besides CP.
    Another strength of our framework is the simplicity of the interface; similar to SAT Modulo Theories architecture, the interface could be enriched to investigate the impact of additional information on the DP solver.
    Another direction is to explore what is represented in the CP model, e.g., the DP model could be a relaxation, while the CP model contains the absent elements of the DP model.
    Finally, we experimented with reducing propagation overhead but obtained inconclusive results, warranting further work into, e.g., reducing redundant work introduced by jumping between states, or by determining \emph{when} to propagate.

\section*{Acknowledgements}
   Imko Marijnissen is supported by the NWO/OCW, as part of the Quantum Software Consortium programme (project number 024.003.037 / 3368).\\ 
   Ryo Kuroiwa is supported by JSPS KAKENHI grant number JP25K24378.

\bibliography{aaai2026}

\potentialremove{
\appendix

\section*{Appendices}

\section{Optimality Gap Analysis}
\label{sec:opt_gap}
    We discuss the impact of adding constraint propagation on the average optimality gap.

    For a minimisation problem, given a lower-bound $Dual$ and an upper-bound $Primal$ on the objective, Equation~\eqref{eq:opt_gap} defines the \textit{optimality gap}.
    The optimality gap provides information about how far the lower- and upper-bound are from each other, the closer, the better.
    \begin{equation}
        \label{eq:opt_gap}
        Optimality Gap = \frac{Primal - Dual}{\max\{1, Primal\}}
    \end{equation}
    We calculate the average optimality gap by starting with $1.0$ for every instance for which infeasibility was not proven, and then updating the optimality gap for an instance whenever a new dual or primal bound is found, taking the average of all instances as the average optimality gap. 
    We do not plot \astar{} since it does not find intermediate solutions. 

    \subsection{$\bm{\wttw}$}
    \label{sec:wttw_opt_gap}
        Figure~\ref{fig:wttw_opt_gap} shows that, besides being able to solve more instances (as shown in Figure~\ref{fig:wttw_overview}), our approach also reaches a better optimality gap in fewer states.
        This indicates that pruning also results in better bounds/solutions.

        Furthermore, Figure~\ref{fig:wttw_opt_gap_time} shows that the optimality gap between CABS and CABS+CP over time is very similar.
        While our approach guides the search better and is able to solve more instances, the overhead of propagation causes a similarity in the optimality gap between the two approaches.

            \begin{figure}[hbtp]
                \begin{subfigure}[htbp]{\columnwidth}
                    \includegraphics[scale=1.0]{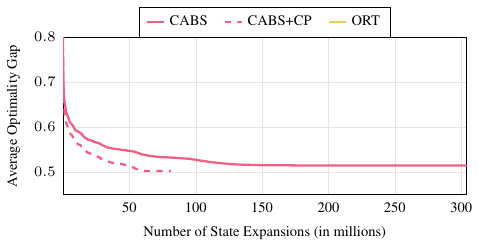}
                    \caption{Average optimality gap over number of state expansions for \wttw{} instances. Our approach achieves a smaller optimality gap using fewer states compared to CABS.}
                    \label{fig:wttw_opt_gap}
                \end{subfigure}

                \begin{subfigure}[htbp]{\columnwidth}
                    \includegraphics[scale=1.0]{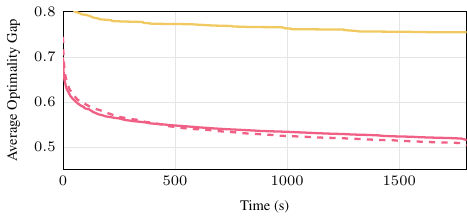}
                    \caption{Average optimality gap over time for \wttw{} instances. Our approach achieves a better optimality gap than CABS.}
                    \label{fig:wttw_opt_gap_time}
                \end{subfigure}
                \caption{Average optimality gap over the number of state expansions and time for \wttw{} instances.}
            \end{figure}

    \subsection{RCPSP}
    \label{sec:rcpsp_opt_gap}
        For RCPSP, Figure~\ref{fig:rcpsp_opt_gap} shows that CABS(+CP) quickly approach an optimality gap of zero, indicating that it is the proving of optimality which is difficult for the approaches.
        Nevertheless, in addition to our approach substantially increasing the number of solved instances (Figure~\ref{fig:rcpsp_overview}), our approach also reduces the optimality gap compared to CABS.
        These results further exhibit the importance of constraint propagation for DP-based approaches when solving RCPSP.

        Looking at the optimality gap over time, Figure~\ref{fig:rcpsp_opt_gap_time} shows a similar trend to Figure~\ref{fig:rcpsp_opt_gap}, exhibiting that the optimality gap is also reduced over time by the addition of constraint propagation.
        Interestingly, both CABS and CABS+CP achieve a lower optimality gap than OR-Tools, showing the value of DP-based approaches for RCPSP.
            \begin{figure}[hbtp]
                \begin{subfigure}[htbp]{\columnwidth}
                    \includegraphics[scale=1.0]{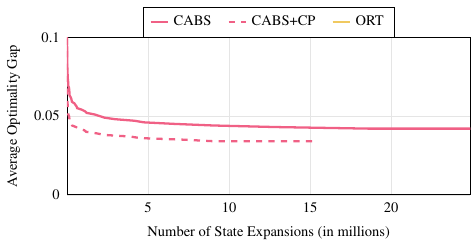}
                    \caption{Average optimality gap over number of state expansions for RCPSP. Our approach achieves a slightly better optimality gap than CABS, indicating that proving optimality is the difficulty.}
                    \label{fig:rcpsp_opt_gap}
                \end{subfigure}
                \begin{subfigure}[htbp]{\columnwidth}
                    \includegraphics[scale=1.0]{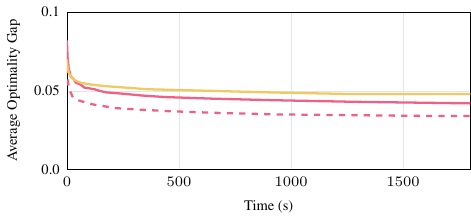}
                    \caption{Average optimality gap over time for RCPSP instances. Our approach achieves a better optimality gap than CABS and OR-Tools.}
                    \label{fig:rcpsp_opt_gap_time}
                \end{subfigure}
                \caption{Average optimality gap over the number of state expansions and time for RCPSP instances.}
            \end{figure}

            This conclusion is further strengthened by Figure~\ref{fig:rcpsp_opt_gap_comparison}, which shows that, for 70 instances, our approach outperforms OR-Tools in terms of optimality gap.
            \begin{figure}[hbtp]
                \centering
                \includegraphics[scale=1.0]{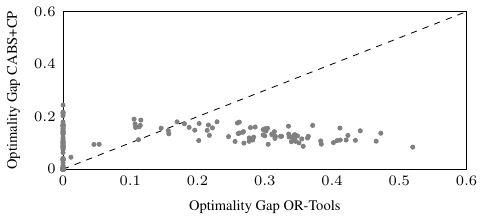}
                \caption{Comparison of optimality gap on RCPSP instances. Our approach improves the optimality gap of OR-Tools for numerous instances.}
                \label{fig:rcpsp_opt_gap_comparison}
        \end{figure}

    \subsection{TSPTW}
    \label{sec:tsptw_opt_gap}
        Similar to the conclusions of Section~\ref{sec:tsptw}, Figure~\ref{fig:tsptw_opt_gap} shows that adding constraint propagation does not improve the optimality gap for the general instances.
        Since our approach times out for many instances, it reaches a higher optimality gap than CABS.

        The effect of propagation overhead with weak pruning can also be seen in Figure~\ref{fig:tsptw_opt_gap_time}, showing that the propagation causes a slower convergence.

        \paragraph{Parameter Analysis}
            Looking at the tightly constrained instances from Section~\ref{sec:tsptw}, Figure~\ref{fig:tsptw_specific_opt_gap} shows that our approach reaches a similar optimality gap as CABS, confirming the findings of Figure~\ref{fig:tsptw_specific_comparison} that our approach mostly reduces the number of states on infeasible instances (which are not shown in these plots).

            Similar to Appendix~\ref{sec:wttw_opt_gap}, Figure~\ref{fig:tsptw_specific_opt_gap_time} indicates that future work into reducing the propagation overhead is warranted.
        
        \begin{figure}[htbp]
            \begin{subfigure}[htbp]{\columnwidth}
                \includegraphics[scale=1.0]{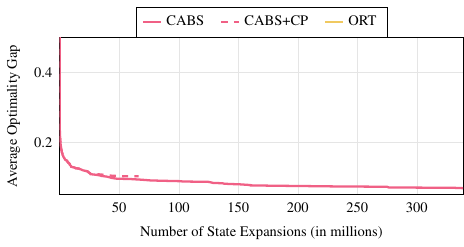}
                \caption{Average optimality gap over number of state expansions for TSPTW instances. Our approach does not improve the optimality gap and times out with fewer state expansions than CABS.}
                \label{fig:tsptw_opt_gap}
            \end{subfigure}   
            \begin{subfigure}[htbp]{\columnwidth}
                \includegraphics[scale=1.0]{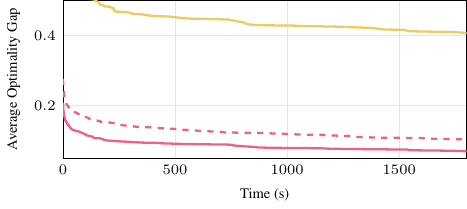}
                \caption{Average optimality gap over time. Our approach does not improve the optimality gap compared to CABS for TSPTW instances.}
                \label{fig:tsptw_opt_gap_time}
            \end{subfigure}

            \caption{Average optimality gap over the number of state expansions and time for TSPTW instances.}
        \end{figure}

        \begin{figure}[htbp]
            \begin{subfigure}[htbp]{\columnwidth}
                \includegraphics[scale=1.0]{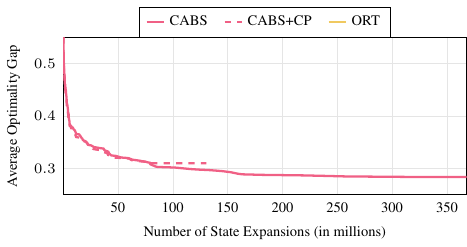}
                \caption{Average optimality gap over number of state expansions for TSPTW instances by \citeauthorandyear{DBLP:journals/jair/RifkiS25} where $1.0 \leq \alpha \leq 1.5$. Our approach achieves the same optimality gap as CABS.}
                \label{fig:tsptw_specific_opt_gap}
            \end{subfigure}
            \begin{subfigure}[htbp]{\columnwidth}
                \includegraphics[scale=1.0]{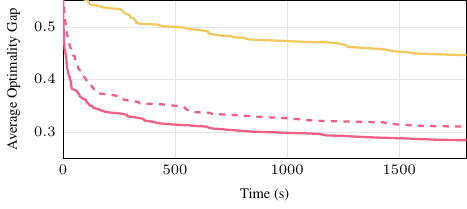}
                \caption{Average optimality gap over time for TSPTW instances by \citeauthorandyear{DBLP:journals/jair/RifkiS25} where $1.0 \leq \alpha \leq 1.5$. CABS reaches the lowest optimality gap, closely followed by our approach.}
                \label{fig:tsptw_specific_opt_gap_time}
            \end{subfigure}
            \caption{Average optimality gap over the number of state expansions and time for TSPTW instances by \citeauthorandyear{DBLP:journals/jair/RifkiS25} where $1.0 \leq \alpha \leq 1.5$.}
        \end{figure}

\section{CP Models}
\label{sec:cp_models}
We describe the models used for OR-Tools in the experiments.
The MiniZinc models can be found in the provided experimental data.

    \subsection{$\bm{\wttw}$}
        The model defines the variable $s_i$ for the start time of job $i$ in Definition~\eqref{eq:cp_wttw_start}.
        Constraint~\eqref{eq:cp_wttw_disjunctive} ensures that the tasks do not overlap.
        Finally, the objective function (Objective~\eqref{eq:cp_wttw_objective}) specifies that the sum of weighted tardinesses is minimised.
        \begin{flalign}
            & \min \sum_{i \in \jobs} \max\{0, s_i + p_i - d_i\} \times w_i & \label{eq:cp_wttw_objective}\\
            & \disjunctive([s_i\ |\ i \in \jobs], [p_i\ |\ i \in \jobs]) & \label{eq:cp_wttw_disjunctive}\\
            & s_i \in [r_i, \delta_i - p_i] & \forall i \in \jobs\label{eq:cp_wttw_start}
        \end{flalign}

    \subsection{RCPSP}
        The model is adapted from the MiniZinc benchmarks repository.

        The model defines two variables: $s_i$ for the start time of task $i$ (Definition~\eqref{eq:cp_rcpsp_start_time}) and the objective variable $o$ (Definition~\eqref{eq:cp_rcpsp_makespan}) representing the makespan.
        
        The constraints then consist of 1) Constraint~\eqref{eq:cp_rcpsp_makespan_cons}, constraining the makespan to be after all of the variables (where $suc_i$ are the successors of task $i$ according to the precedences), 2) Constraint~\eqref{eq:cp_rcpsp_precedences}, constraining the precedences to be respected, and 3) Constraint~\eqref{eq:cp_rcpsp_cumulative}, ensuring that the resource capacities are respected (where $\tasks_r = \{i\ |\ i \in \tasks : u_{ir} > 0 \land p_i > 0\}$, $St_r = [s_i\ |\ i \in \tasks_r]$, $P_r = [p_i\ |\ i \in \tasks_r]$, and $U_r = [u_{ir}\ |\ i \in \tasks_r]$).

        Finally, the objective function (Objective~\eqref{eq:cp_rcpsp_objective}) specifies that the makespan variable is minimised.

        \begin{flalign}
            & \min o & \label{eq:cp_rcpsp_objective}\\
            & s_i + p_i \leq o & \forall i \in \tasks : suc_i \neq \emptyset\label{eq:cp_rcpsp_makespan_cons}\\
            & s_i + p_i \leq s_j & \forall (i, j) \in \precedences\label{eq:cp_rcpsp_precedences}\\
            & \cumulative(St_r, P_r, U_r, C_r) & \forall r \in \resources\label{eq:cp_rcpsp_cumulative}\\
            & s_i \in [0, H] & \forall i \in \tasks \label{eq:cp_rcpsp_start_time}\\
            & o \in [0, H] & \label{eq:cp_rcpsp_makespan}
        \end{flalign}

    \subsection{TSPTW}
        We adapt the model from the 2025 MiniZinc challenge.

        The model defines six variables: 
        1) Definition~\eqref{eq:cp_tsptw_pred}, which defines $pred_i$, representing the predecessor for each location $i$ (i.e. the location which was visited prior to this one), 
        2) Definition~\eqref{eq:cp_tsptw_dur_to_pred}, which defines $durToPred_i$, which represents the travel time from the predecessor of location $i$ to $i$, 
        3) Definition~\eqref{eq:cp_tsptw_arrival}, which defines $arr_i$, representing the arrival time at location $i$,
        4) Definition~\eqref{eq:cp_tsptw_departure}, which defines $dep_i$, representing the departure time at location $i$,
        5) Definition~\eqref{eq:cp_tsptw_departure_pred}, which defines $depPred_i$, representing the departure from the predecessor of location $i$, and
        6) Definition~\eqref{eq:cp_tsptw_objective}, which defines the objective $o$, representing the sum of travel times.

        The following constraints are then specified:
        1) Constraint~\eqref{eq:cp_tsptw_objective_constraint} constrains $o$ to be equal to the sum of travel times,
        2) Constraint~\eqref{eq:cp_tsptw_arr_constraint} constrains $arr_i$ to be equal to the departure time at the predecessor of $i$ plus the duration from the predecessor of $i$ to $i$,
        3) Constraint~\eqref{eq:cp_tsptw_dur_to_pred_constraint} ensures that $durToPred_i$ is equal to the travel time from the predecessor of location $i$ to $i$,
        4) Constraint~\eqref{eq:cp_tsptw_dep_constraint} ensures that the latest possible visiting times are respected and that the departure at location $i$ either occurs at its arrival or at its earliest possible release time,
        5) Constraint~\eqref{eq:cp_tsptw_dep_pred_constraint} ensures that the $depPred_i$ is properly channelled to be equal to the value of the variable $dep_{pred_i}$, and
        6) Constraint~\eqref{eq:cp_tsptw_circuit} constrains the sequence of locations to be a tour using the \constraint{Circuit} constraint.

        Finally, the objective function (Objective~\eqref{eq:cp_tsptw_objective_function}) specifies that the sum of travel times variable is minimised.

        \begin{flalign}
            & \min o & \label{eq:cp_tsptw_objective_function}\\
            & o = \sum_{i \in locations} durToPred_i & \label{eq:cp_tsptw_objective_constraint}\\
            & arr_i = depPred_i + durToPred_i & \forall i \in \locations\label{eq:cp_tsptw_arr_constraint}\\
            & durToPred_i = \distance{pred_i}{i} & \forall i \in \locations\label{eq:cp_tsptw_dur_to_pred_constraint}\\
            & dep_i = \max\{arr_i, r_i\} \leq \delta_i & \forall i \in \locations\label{eq:cp_tsptw_dep_constraint}\\
            & depPred_i = dep_{pred_i} & \forall i \in \locations\label{eq:cp_tsptw_dep_pred_constraint}\\
            & \constraint{Circuit}([pred_i\ |\ \forall i \in \locations]) & \label{eq:cp_tsptw_circuit}\\
            & pred_i \in [0, |\locations|] & \forall i \in \locations\label{eq:cp_tsptw_pred}\\
            & durToPred \in [0, \max_{i, j \in \locations} \distance{i}{j}] & \forall i \in \locations\label{eq:cp_tsptw_dur_to_pred}\\
            & arr_i \in [0, \max_{i \in \locations} \delta_i] & \forall i \in \locations\label{eq:cp_tsptw_arrival}\\
            & dep_i \in [\min_{i \in \locations} r_i, \max_{i \in \locations} \delta_i] & \forall i \in \locations\label{eq:cp_tsptw_departure}\\
            & depPred_i \in [0, \max_{i \in \locations} \delta_i] & \forall i \in \locations\label{eq:cp_tsptw_departure_pred}\\
            & o \in [0, |\locations| \times \max_{i, j \in \locations} \distance{i}{j}] \label{eq:cp_tsptw_objective}
        \end{flalign}
}{}
\end{document}